\documentclass[letterpaper]{article} % DO NOT CHANGE THIS
\usepackage{aaai2026}  % DO NOT CHANGE THIS
\usepackage{times}  % DO NOT CHANGE THIS
\usepackage{helvet}  % DO NOT CHANGE THIS
\usepackage{courier}  % DO NOT CHANGE THIS
\usepackage[hyphens]{url}  % DO NOT CHANGE THIS
\usepackage{graphicx} % DO NOT CHANGE THIS
\usepackage{natbib}  % DO NOT CHANGE THIS AND DO NOT ADD ANY OPTIONS TO IT
\usepackage{caption} % DO NOT CHANGE THIS AND DO NOT ADD ANY OPTIONS TO IT
\usepackage{algorithm}
\usepackage{algorithmic}
\usepackage{amssymb}
\usepackage{enumitem}
\usepackage{amsmath}
\usepackage{booktabs}
\usepackage{multirow}
\usepackage{newfloat}
\usepackage{listings}
\DeclareCaptionStyle{ruled}{labelfont=normalfont,labelsep=colon,strut=off} % DO NOT CHANGE THIS
\floatstyle{ruled}
\newfloat{listing}{tb}{lst}{}
\floatname{listing}{Listing}
\title{DarkFarseer: Robust Spatio-temporal Kriging under Graph Sparsity and Noise}
\author{
    Zhuoxuan Liang\textsuperscript{\rm 1},
    Wei Li\thanks{Corresponding author}\textsuperscript{\rm 1,2},
    Dalin Zhang\textsuperscript{\rm 3},
    Ziyu Jia\textsuperscript{\rm 4,5},
    Yidan Chen\textsuperscript{\rm 1},
    Zhihong Wang\textsuperscript{\rm 1},
    Xiangping Zheng\textsuperscript{\rm 1,2},
    Moustafa Youssef\textsuperscript{\rm 6}\\
}
\affiliations{
    \textsuperscript{\rm 1} College of Computer Science and Technology, Harbin Engineering University, Harbin, China\\
    \textsuperscript{\rm 2} Modeling and Emulation in E-Government National Engineering Laboratory, Harbin Engineering University, Harbin, China\\
    \textsuperscript{\rm 3} Space Information Research Institute, Hangzhou Dianzi University, Hangzhou, China\\
    \textsuperscript{\rm 4} Institute of Automation, Chinese Academy of Sciences, Beijing, China.\\
    \textsuperscript{\rm 5} Shanghai Key Laboratory of Data Science, Shanghai, China.\\
    \textsuperscript{\rm 6} Egyptian Wireless Research Center of Excellence, The American University in Cairo, Cairo, Egypt
    \{zz.liang, wei.li, chenyidan, 1084472397, xpzheng\}@hrbeu.edu.cn, zhangdalin@hdu.edu.cn, jia.ziyu@outlook.com, moustafa-youssef@aucegypt.edu
}

\usepackage{bibentry}
\begin{document}

\maketitle

\begin{abstract}
The rapid expansion of the Internet of Things (IoT) has created a growing demand for large-scale sensor deployment. However, the high cost of physical sensors limits the scalability and coverage of sensor networks, making fine-grained sensing difficult. Inductive Spatio-Temporal Kriging (ISK) addresses this challenge by introducing virtual sensors that infer measurements from physical sensors, typically using graph neural networks (GNNs) to model their relationships.
Despite its promise, current ISK methods often rely on standard message-passing and generic architectures that fail to effectively capture spatio-temporal features or represent virtual nodes accurately. Additionally, existing graph construction techniques suffer from sparse and noisy connections, further hindering performance.
To address these limitations, we propose DarkFarseer, a novel ISK framework with three key innovations. First, the Style-enhanced Temporal-Spatial architecture adopts a temporal-then-spatial processing scheme with a temporal style transfer mechanism to enhance virtual node representations. Second, Regional-semantic Contrastive Learning improves representation learning by aligning virtual nodes with regional component patterns. Third, the Similarity-Based Graph Denoising Strategy mitigates the influence of noisy edges by leveraging temporal similarity and regional structure.
Extensive experiments on real-world datasets demonstrate that DarkFarseer significantly outperforms state-of-the-art ISK methods.
\end{abstract}

\section{Introduction}\label{sct:section_1}
The rapid advancement and widespread adoption of the Internet of Things (IoT) have significantly increased the demand for fine-grained and ubiquitous sensing. As a result, the large-scale deployment of sensors has become a common objective across various domains~\cite{zhou2021informer, liu2022scinet, belavadi2020air}. However, the high cost associated with sensor deployment presents a substantial challenge in this regard.
To mitigate this limitation, Spatio-Temporal Kriging (SK) has emerged as a promising solution. SK introduces the concept of virtual sensors, aiming to estimate their readings based on data collected from physically deployed sensors. Existing SK methods are generally categorized into two branches: Transductive SK (TSK) and Inductive SK (ISK). TSK assumes prior knowledge of virtual sensor locations during training, whereas ISK infers virtual sensor readings without such information until inference time, making it more practical and scalable for real-world applications.
Building on prior work in SK~\cite{wu2021inductive, xu2023kits, zheng2023increase, li2024non}, this paper focuses on the ISK task to further explore its potential and improve its effectiveness.

Recent deep learning-based approaches~\cite{appleby2020kriging, wu2021inductive, hu2023decoupling, xu2023kits, li2024non, wei2024inductive}, which represent the state of the art (SOTA) in inductive spatio-temporal kriging (ISK), predominantly leverage Graph Neural Networks (GNNs) to model the relationships between physical and virtual sensors—conceptualized as nodes in a graph. Despite their success, most existing methods adopt generic GNN architectures without considering the unique characteristics and challenges inherent to the ISK task. This oversight results in suboptimal graph construction and network design, ultimately limiting model performance.

\textbf{L1: Limitations of spatial-first structures in modeling spatial relationships.} Existing GNN-based ISK models adopt a spatial-first paradigm, where spatial dependencies are modeled prior to temporal encoding. However, this approach computes node embeddings directly from raw time series data without adequate temporal abstraction, weakening the effectiveness of spatial message passing, which is a critical component for accurate virtual sensor inference in ISK tasks.

\textbf{L2: Lack of specific message-passing design in GNNs for virtual nodes.} Existing GNN-based ISK methods~\cite{wu2021inductive, hu2023decoupling, xu2023kits, li2024non} typically adopt generic message-passing frameworks that treat all nodes uniformly, aiming to capture global spatial information across the graph. These models aggregate features from multi-hop neighbors to learn node representations and derive graph-level embeddings. However, ISK has a specific goal: accurately inferring the values of virtual nodes. This objective requires emphasizing the contribution of immediate neighbors rather than relying on information from the broader graph. Furthermore, simple weighted aggregation fails to capture the complex interactions between virtual nodes and their physical neighbors. Therefore, a task-specific message-passing strategy tailored to virtual node inference is crucial for improving ISK performance.

\textbf{L3: Challenges from sparsity and noise in graph structures.} Statistical analysis and preliminary experiments with a representative GNN-based ISK method~\cite{wu2021inductive} across multiple datasets reveal two major issues: edge sparsity in expert-defined graphs and edge noise in proximity-based graphs. Pairwise Connectivity Graphs (PCGs), built via expert knowledge, often suffer from severe sparsity, limiting information flow from neighbors and hindering virtual node inference (see Appendix \textbf{A.2.1}). In contrast, Spatial Proximity Graphs (SPGs) exhibit high edge density, which introduces noise by connecting weakly relevant neighbors, thus degrading embedding quality (see Appendix \textbf{A.2.2}). Both sparse and noisy graph structures adversely affect ISK performance and must be carefully addressed.

Considering these limitations, we first conduct a comprehensive empirical study to demonstrate the sparsity and noise issues (\textbf{L3}) in existing graph structures. To address these challenges, we propose DarkFarseer, a novel framework that mitigates these limitations and achieves new SOTA performance on ISK tasks.

To address \textbf{L1} and \textbf{L2}, we propose the Style-enhanced Temporal-Spatial (\textsc{SeTS}) architecture. Unlike existing methods~\cite{hu2023decoupling}, which adopt a spatial-then-temporal pipeline, \textsc{SeTS} follows a temporal-then-spatial design, where temporal features are first extracted and then used for spatial modeling.
For temporal feature extraction, we adopt the channel-independence strategy~\cite{zeng2023transformers}, which independently captures temporal fluctuations for each neighbor of a virtual node. For spatial message passing, we focus on the 1-hop neighbors of virtual nodes and apply a style transfer mechanism~\cite{liu2024zepo} that transfers the temporal fluctuation patterns of physical neighbors to virtual nodes, based on temporal similarity. 
The use of 1-hop neighbors ensures a balance between effectiveness and computational efficiency.

To address the \textbf{sparsity issue in L3}, we propose the Regional-semantic Contrastive Learning (\textsc{RsCL}) approach, which leverages regional information as contrastive samples rather than relying solely on individual nodes. This enriches the representations of nodes with sparse spatial links and alleviates the impact of sparsity. Specifically, we identify highly connected subgraphs, or Biconnected Components (BCCs), and aggregate them into prototypes~\cite{ma2023graph, li2021prototypical} to capture high-level regional semantics. Within a contrastive learning framework, \textsc{RsCL} encourages virtual nodes to align with their associated prototypes while pushing them away from unrelated ones, thereby enhancing their alignment with regional patterns.
Moreover, unlike existing ISK CL methods~\cite{li2024non} that treat all nodes as anchors, \textsc{RsCL} uses only virtual nodes as anchors and performs fine-grained contrast, leading to more discriminative representations for sparsely connected virtual nodes.
To address the \textbf{noise issue in L3}, we introduce the Similarity-based Graph Denoising Strategy (\textsc{SGDs}), which reduces the influence of noisy edges. \textsc{SGDs} evaluates both the temporal similarity between each virtual node and its neighbors and their similarity to regional prototypes. It then selectively downweights the connections to dissimilar neighbors, refining the graph structure and improving representation quality.

The contributions of this paper are as follows:
\begin{itemize}[leftmargin=*]
    \item We conduct an in-depth analysis of existing graph construction methods and, through preliminary experiments, reveal two limitations: sparsity and noise in graph structures that limit the effectiveness of ISK tasks.
    
    \item We propose the \textsc{SeTS} architecture, which adopts a temporal-then-spatial scheme and incorporates a temporal style transfer mechanism, enhancing virtual node representations by effectively capturing and fusing the fluctuation characteristics of their neighbors.
    
    \item We introduce the \textsc{RsCL} strategy and the \textsc{SGDs} module to explicitly address graph sparsity and noise. These components enhance virtual node learning by leveraging region-aware contrastive learning and edge filtering.
    
    \item We evaluate the proposed framework on multiple real-world datasets, comparing it with SOTA baselines. The results demonstrate significant performance improvements.
\end{itemize}

\section{Problem Definition}\label{sct:section_2}
\subsection{Spatio-temporal Kriging (SK)}
Given the observations \( \mathbf{X}^o_{t:t+T} \in \mathbb{R}^{N_o \times T} \) from \( N_o \) physical sensors located at \( \mathcal{L}^o = \{l_i^o\}_{i=1}^{N_o} \), the objective of SK is to infer the readings \( \mathbf{X}^u_{t:t+T} \in \mathbb{R}^{N_u \times T} \) for \( N_u \) virtual sensors at locations \( \mathcal{L}^u = \{l_i^u\}_{i=1}^{N_u} \). These virtual sensors are spatially related to the physical ones through spatial relationships (e.g., geographical distance). 
The data \( \mathbf{X}^o_{t:t+T} \) and \( \mathbf{X}^u_{t:t+T} \) represent multivariate time series for the observed and virtual sensors, respectively. 
The spatial relationships between the physical and virtual sensors are captured in a set of triplets \( \mathcal{R} = \{ (l_i^o, l_j^u, d_{i,j}) \mid l_i^o \in \mathcal{L}^o, l_j^u \in \mathcal{L}^u \} \), where \( d_{i,j} \) denotes the spatial distance between the physical sensor \( l_i^o \) and the virtual sensor \( l_j^u \). These spatial relationships form the foundation for modeling the inference process in SK.

\paragraph{\textbf{Inductive SK (ISK)}.}
In ISK, the model only has access to the spatial relationships between physical nodes during the training phase, but during the inference phase, it can access the complete spatial relationships between both physical and virtual nodes. Let \( \mathcal{F} \) denote the kriging model. The kriging process during training can be formulated as:
\begin{equation}
    \hat{\mathbf{X}}_{t:t+p}^o = \mathcal{F}_{\text{train}}(\mathbf{X}^o_{t:t+p}, \mathcal{R}^o), 
\end{equation}
where \( p \) represents the time window size. \( \mathcal{R}^o = \{ (l_i^o, l_j^o, d_{i,j}^o) \mid l_i^o, l_j^o \in \mathcal{L}^o, i \neq j \} \) represents the spatial relationships of the observed sensors. The inference phase for ISK can be formulated as:
\begin{equation}
    \hat{\mathbf{X}}_{t:t+p}^u = \mathcal{F}_{\text{infer}}(\mathbf{X}^o_{t:t+p}, \mathcal{R}),
\end{equation}
where \( \mathcal{R} \) represents the complete spatial relationships between both observed and virtual nodes.

\subsubsection{Graph-based ISK.}
By considering sensors as nodes $\mathcal{V}$ and the reachability relationships between nodes as edges $\mathcal{E}$, sensors and their spatial relationships can be represented as a graph \( \mathcal{G} = (\mathcal{V}, \mathcal{E}) \).
The spatial relationships $R$ in ISK can be represented using an adjacency matrix \( \textbf{A} \in \mathbb{R}^{N \times N} \). On top of this, the training and inference phases of graph-based ISK can be expressed as:
\begin{equation}
	\hat{\textbf{X}}_{t:t+p}^o=\mathcal{F}_{\text{train}}(\textbf{X}^o_{t:t+p},\; \textbf{A}^o), \text{where}\;\textbf{A}^o\in \mathbb{R}^{N_o\times N_o},
\end{equation}
\begin{equation}
	\text{and}\; \hat{\textbf{X}}_{t:t+p}^u=\mathcal{F}_{\text{infer}}(\textbf{X}^o_{t:t+p}, \textbf{A}),\; \text{where}\;\textbf{A}\in \mathbb{R}^{N\times N}.
\end{equation}

\section{Analysis and Motivation}\label{sct:section_3}
In this section, we showcase the sparsity and noise issues in current graph structures. We first investigate two types of graph structures. Subsequently, we analyze the issues of inconsistent graph density (Appendix \textbf{A.2.1}) and noisy edges (Appendix \textbf{A.2.2}) present in the two graph construction methods. Finally, we obtain three observations.

\subsection{Graphs for ISK}\label{sct:section_3_1}
Deep learning-based kriging methods primarily rely on GNNs that necessitate constructing a graph to model the interrelationships of sensors (i.e., nodes in a graph). 
Heuristic-based Graphs~\cite{jin2024survey} based on the distances between nodes is a common and effective approach~\cite{wu2019graph, song2020spatial, li2024gpt}. There are generally two branches to constructing heuristic-based Graphs:

\textit{Pairwise Connectivity Graphs (PCGs)}: In this approach, the reachability and corresponding distance between two nodes are available according to expert intervention. Thus, the adjacency matrix can be built based on the pairwise distance:
\begin{equation}
    \textbf{A}_{i,j} = \left\{\begin{matrix}
        d_{i,j} &  \text{if node $i$ and $j$ are linked},\\
        0 &  \text{otherwise},\\
    \end{matrix}\right.
\end{equation}
where $d_{i,j}$ is the distance between node $i$ and $j$.
	
\textit{Spatial Proximity Graphs (SPGs)}: The reachability between nodes is unknown while the geographical coordinates of each node are available. Thus, it is essential to calculate the distance between any two nodes to construct a graph. The adjacency matrix can be formulated as:
\begin{equation}
    \label{eq:SPGs}
    \textbf{A}_{i,j} = \left\{\begin{matrix}
        \frac{\text{exp}(-||d_{i,j}||_2)}{\sigma} &  \text{if $\frac{\text{exp}(-||d_{i,j}||_2)}{\sigma} < \varepsilon$},\\
        0 &  \text{otherwise}.\\
    \end{matrix}\right.
\end{equation}
where $d_{i,j}$ is the node distance, $\sigma$ is a hyper-parameter to normalize the distance distribution, and \(\varepsilon\) is the threshold to control the sparsity of the adjacency matrix. 

Notably, PCGs require expert intervention for pairwise node distances, while SPGs connect nodes based on geographic distances and a pre-defined threshold \(\varepsilon\).

\subsection{Motivations}\label{sct:section_3_2}
We conducted comprehensive preliminary experiments and made observations to confirm our conjectures and motivate the proposal of DarkFarseer. 
We present the results and analysis in Appendix \textbf{A.}, and only report three key observations as follows:

\textbf{Observation I:} The PCGs are constrained by sparsity, which hinders their performance. However, these limitations can be alleviated by incorporating additional topological prompts into the ISK model.

\textbf{Observation II:} The PCGs are subject to a density limitation: excessive edges introduce noise into the model. Thus, designing noise-reduction strategies can help mitigate this issue and enhance performance.

\textbf{Observation III:} Although increasing $\epsilon$ to make the SPGs sparser can help, noise caused by edges constructed from geographical distance still persists. In other words, the noise edges inherent to SPGs remain, regardless of the selected value for $\epsilon$.

Considering these observations, it is imperative to develop a framework that can mitigate the shortcomings associated with both PCGs and SPGs. This framework should possess two key features: (1) The capability to employ topological prompts to counteract potential graph sparsity. (2) A noise reduction strategy designed to diminish noise resulting from either excessive graph density or misleading information intrinsic to the edges themselves.
\begin{figure*}[htbp]
	\centering
	\includegraphics[width=0.9\textwidth]
    {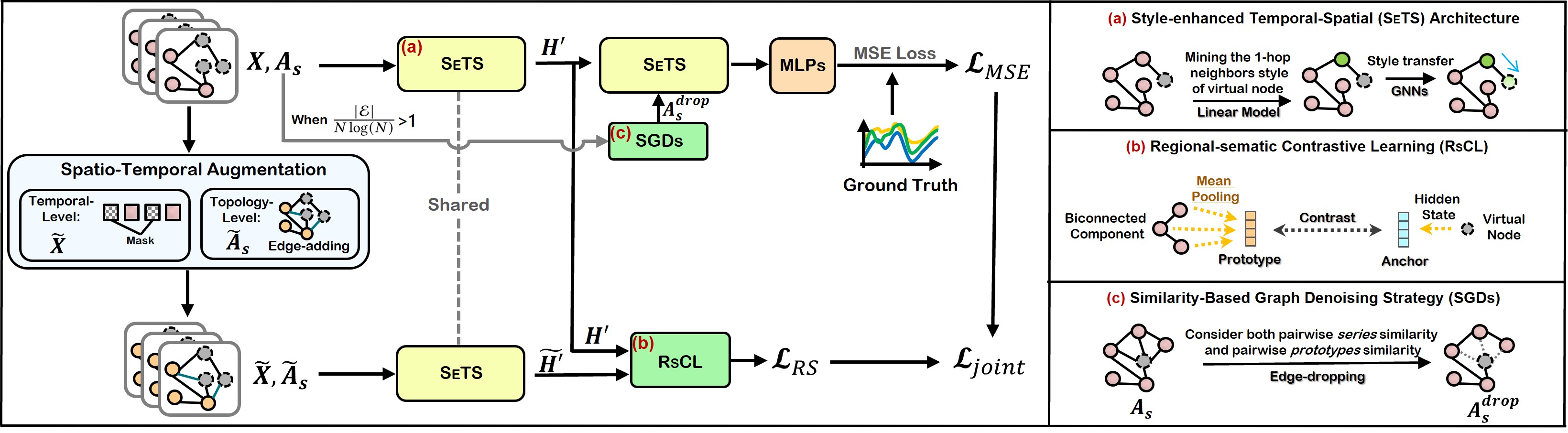}
		% \vspace{-1.5mm}
	\caption{Overall architecture of DarkFarseer.
    }
	% \vspace{-5mm}
	\label{fig:framework}
\end{figure*}
\section{Methodology}\label{sct:section_4}
\subsection{DarkFarseer Overview}
\subsubsection{Training Protocol.}\label{sct:section_4_1_1}
The decrement strategy~\cite{wu2021inductive} is the prevailing training approach for ISK. In line with this, we adopt the decrement training strategy as proposed by IGNNK~\cite{wu2021inductive}. Specifically, during the training phase, we randomly select \(N_u\) nodes to serve as the virtual nodes to be predicted, while using the subgraph formed by the remaining \(N_o - N_u\) nodes to optimize the kriging model. 

\subsubsection{Architecture.}
The overall architecture of DarkFarseer is illustrated in Figure \ref{fig:framework}. At its core, DarkFarseer leverages graph structures and observed node sequences to jointly learn patterns for virtual nodes. It consists of three key components: Style-enhanced Temporal-Spatial (\textsc{SeTS}) architecture, Regional-sematic Contrastive Learning (\textsc{RsCL}), and Similarity-Based Graph Denoising Strategy (SGDs).
The \textsc{SeTS} architecture employs a temporal-then-spatial fashion and equips with a fluctuation style transfer strategy to represent virtual nodes, as depicted in Figure \ref{fig:framework}(a).
\textsc{RsCL} operates on two graph views to compute the contrastive loss \(\mathcal{L}_{RS}\). It utilizes graph topological information to create prototypes, emphasizing the relationship between virtual node patterns and regional prototype patterns. This mitigates graph sparsity, as illustrated in Figure \ref{fig:framework}(b).
SGDs reduces graph noise by adjusting edge weights based on the similarity of node sequences and their prototypes. The resulting denoised graph structure, $A_s^{drop}$, is shown in Figure \ref{fig:framework}(c).
Finally, \textsc{SeTS} re-encodes spatio-temporal patterns using the denoised graph. DarkFarseer is optimized with a combination of prediction loss \(\mathcal{L}_{MSE}\) and contrastive loss \(\mathcal{L}_{RS}\).
The subsequent sections provide detailed explanations of each module.

\subsection{Style-enhanced Temporal-Spatial (\textsc{SeTS}) Architecture}\label{sct:section_4_2}
\textsc{SeTS} first employs DLinear to independently encode each node’s seasonality and trend—leveraging channel–independence techniques~\cite{nie2023a, han2024capacity, dong2024simmtm}—so that the temporal dynamics of all observed neighbours are fully represented before any spatial aggregation, thus eliminating \textbf{L1}. Guided by the complete adjacency matrix $\mathbf{A}$, we then perform a fluctuation–style transfer: the DLinear~\cite{zeng2023transformers}‐encoded temporal style of every virtual node’s 1-hop neighbours is propagated to the virtual node itself, injecting missing temporal knowledge to mitigate \textbf{L2}, after which spatial modules capture inter-node interactions.

Let \( \mathcal{N}(i) \) represent the set of neighbors of the virtual node \(i\). To begin, we apply DLinear to extract hidden style sequences \( \textbf{S} \) for each corresponding time series of the neighbors. This operation produces temporal style sequences $S_j$ for each neighbor $j$ of virtual node $i$:
\begin{equation}
        \label{eq:dlinear}
	\textbf{S}_{j} = \sigma(\text{DLinear}(\textbf{X}_{s,j})),\, j\in \mathcal{N}(i), \, i \in \mathcal{V}^u,
\end{equation}
where \( \sigma(\cdot) \) is the Sigmoid activation function. 
Next, we fuse these style sequences with the corresponding time series data of the virtual node neighbors. We perform this fusion via element-wise multiplication:
\begin{equation}
        \label{eq:hadamard}
	\textbf{H}_{j} = \textbf{H}_{j} \odot \textbf{S}_{j},\, j\in \mathcal{N}(i), \, i \in \mathcal{V}^u,
\end{equation}
where \( \odot \) denotes the Hadamard product. Here, \( \textbf{H} \in \mathbb{R}^{(N_o - N_u) \times p \times \varphi} \) represents the hidden state, with \( p \) being the number of time steps and \( \varphi \) the hidden dimension.

We introduce the MPNN~\cite{cini2022filling} as the GNN backbone, which is expressed as:
\begin{equation}
	\textbf{H}' = \text{MPNN}(\textbf{H},\textbf{A}_s),
\end{equation}
where \( \textbf{H}' \) is the updated hidden states after message passing.

\subsection{Regional-sematic Contrastive Learning (\textsc{RsCL})}\label{sct:section_4_3}
Building on \textbf{Observation \uppercase\expandafter{\romannumeral1}}, we design the \textsc{RsCL} to enrich the representation of virtual nodes and address the limitations of sparse graphs.

For ISK tasks, we adopt two augmentation methods: (1) Temporal Augmentation, where each sequence element within a time window is masked with a probability \(p_t=0.2\); and (2) Topology Augmentation, where edges are added between unconnected nodes with a probability \( p_s =0.003\).

CL design involves selecting three types of objects: anchors, positive samples, and negative samples.

\textit{Anchor Selection}.
Anchors in CL highlight target features. In ISK, we set virtual nodes as the only anchors, instead of all nodes as in \cite{li2024non}, which reduces complexity and keeps the model focused on their temporal patterns.

\textit{Biconnected Component (BCC)}.
BCCs are subgraphs with high local dependency, capable of representing regional information~\cite{ma2023rethinking}. 
Following~\cite{ma2023rethinking}, we set a threshold \(\mu\), and only edges with values greater than \(\mu\) are used for solving BCC. In other words, \(\mu\) is used to control the sparsity of the BCC. Subsequently, we use the Tarjan algorithm~\cite{tarjan1972depth} to identify the BCCs of the given graph.
Mathematically, we represent each node $i$ in the graph and its corresponding BCC $C_j$ using $\{ (i, C_j) \mid i \in \mathcal{V}, i \in C_j, j \in \{1, 2, ..., m\} \}$, where $m$ is the number of solved BCCs.

\textit{Positive and Negative Samples Selection}.
To fully leverage topological prompts, we decide to combine Prototypical Contrastive Learning~\cite{li2021prototypical} and BCC.
Inspired by~\cite{ma2023graph, ji2023spatio}, we construct soft cluster centers as prototypes through BCC. Specifically, for each BCC \( C_i \), we calculate the average hidden state of all nodes within \( C_i \) as the prototype, which can be formalized as:
\begin{equation}
	\textbf{P}_i = \frac{1}{|C_i|}\sum_{j \in C_i}^{} \widetilde{\textbf{H}'}_{j},
\end{equation}
where $\widetilde{\textbf{H}}$ is the hidden state of augmented view.
This contrast connects virtual node patterns to regional BCC patterns.
However, constructing this contrast requires considering two scenarios: \textbf{(1)} A virtual node (anchor) exists in two or more BCCs; \textbf{(2)} 
The prototypes of other BCCs corresponding to the remaining nodes in the BCC containing the virtual node cannot be used as negative samples.

\begin{figure}[]
	\centering
	\includegraphics[width=0.9\linewidth]{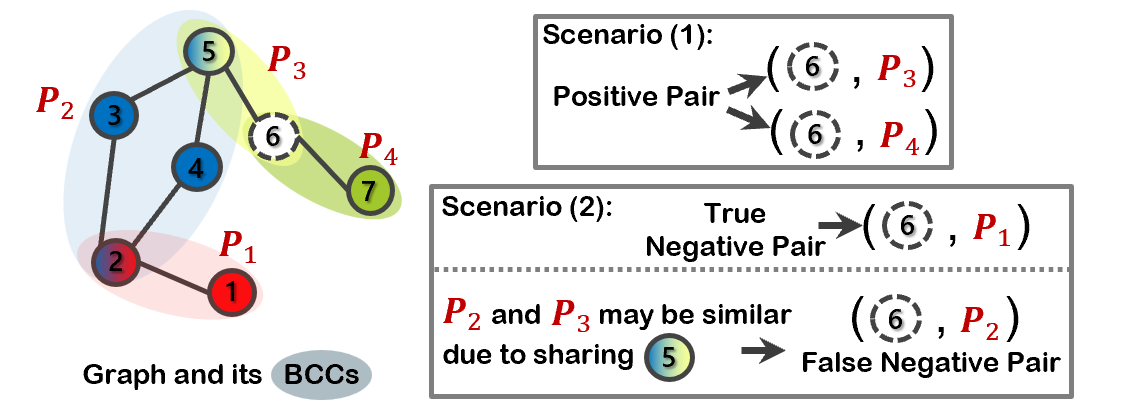}
	\caption{Illustrations of Contrastive Scenario (1) and (2).
    }
	\label{fig:contrast}
\end{figure}

For \textbf{scenario (1)}, as shown in Figure \ref{fig:contrast}, the virtual node 6 appears in both
 \(\textbf{P}_3\) and \(\textbf{P}_4\). \(\textbf{P}_3\) and \(\textbf{P}_4\) should have similar patterns to node 6, so \(\textbf{P}_3\) and \(\textbf{P}_4\) should be considered positive samples. That is, all prototypes corresponding to the BCCs where the virtual node is located should be considered \textbf{positive samples}.

For \textbf{scenario (2)}, in Figure \ref{fig:contrast}, the virtual node 6 is located in the BCC corresponding to prototype \(\textbf{P}_3\) (yellow area). This BCC includes node 5, which is also located in the BCC corresponding to \(\textbf{P}_2\) (blue area). This means node 5 shares part of the pattern of \(\textbf{P}_3\) and part of the pattern of \(\textbf{P}_2\). Thus, \(\textbf{P}_2\) would become a false negative sample. Therefore, when selecting negative samples, we filter out these types of prototypes to avoid generating \textit{false negative samples}. On the other hand, \(\textbf{P}_1\) does not share any nodes with \(\textbf{P}_3\) or \(\textbf{P}_4\), so \(\textbf{P}_1\) can be used as a \textbf{negative sample}. 

Considering the above two scenarios,
the InfoNCE~\cite{oord2018representation} for a single virtual node \(i\) can be expressed as:
\begin{equation}
\label{eq:cl_loss}
\resizebox{1\linewidth}{!}{$
	CL(h_i) = \frac{1}{|P_i^+|} \sum_{\textbf{P}_k\in P^+_i}^{} -log \frac{\text{exp}(\text{s}(h_i, \textbf{P}_k)/\tau)}{\text{exp}(\text{s}(h_i, \textbf{P}_k)/\tau)+\sum_{\textbf{P}_j\in P^-_i}^{}\text{exp}(\text{s}(h_i, \textbf{P}_j)/\tau)}$,
    }
\end{equation}
where $h_i$ is the hidden state of virtual node $i$. \(s(\cdot,\cdot)\) is the cosine similarity function. $\tau$ is the temperature hyper-parameter. \(P_i^+\) is the set of prototypes that should be positive samples (\textbf{scenario (1)}) of virtual node $i$. \(P_i^-\) is the set of prototypes that should be negative samples (\textbf{scenario (2)}) of virtual node $i$. The total contrastive loss that the model needs to optimize is:
\begin{equation}
	\mathcal{L}_{RS} = \frac{1}{|\mathcal{V}^o|} \sum_{i\in\mathcal{V}^o}^{} CL(h_i).
\end{equation}

\subsubsection{Training Objective.}
Unlike previous works~\cite{wu2021inductive} that reconstruct all nodes to train the ISK model, DarkFarseer focuses on virtual nodes by training with a randomly selected sequence of \(\mathcal{V}^u\) nodes, optimizing both MSE (Mean Squared Error) loss and contrastive loss \(\mathcal{L}_{RS}\):
\begin{equation}
	\mathcal{L}_{joint} = \frac{1}{|\mathcal{V}^u|p}\sum_{i\in\mathcal{V}^u}\sum_{t=1}^{p}(x^{o}_{i,t}-\hat{x}_{i,t})^2	+\eta\,\mathcal{L}_{RS},
\end{equation}
where \(\eta\) is the weight for the contrastive loss \(\mathcal{L}_{RS}\).

\subsection{Similarity-based Graph Denoising Strategy (SGDs)}\label{sct:section_4_4}
Combining \textbf{Observation \uppercase\expandafter{\romannumeral2}} and \textbf{Observation \uppercase\expandafter{\romannumeral3}}
, we highlight the need for a noise reduction to address noisy edges in dense PCGs and SPGs. Thus, in this section, we propose a simple yet effective graph denoising strategy for ISK.

GNNs are sensitive to structure noise~\cite{gilmer2017neural}.
To address this, inspired by~\cite{dai2024comprehensive, ju2024survey}
, we propose SGDs, which optimizes graph structures and reduces the impact of noisy edges to improve the robustness of ISK models. Our focus is on edges related to virtual nodes.
Specifically, for each virtual node \(i \in \mathcal{V}^u\), we assess the sequence similarity and prototype similarity between it and each of its neighboring nodes \(j\in \mathcal{N}(i)\), obtaining the weighted similarity \(\gamma_{i,j}\):
\begin{equation}
        \label{eq:gamma}
	\gamma_{i,j} = \alpha \, s(\textbf{P}_i, \textbf{P}_j) + (1-\alpha) \, s(\textbf{H}'_{i}, \textbf{H}'_{j}),
\end{equation}
where \(\alpha=\frac{1}{\mathcal{D}}\) balances the influence of different similarities, with $\mathcal{D}=\frac{|\mathcal{E}|}{N\log (N)}$ measuring graph sparsity. This ensures prototype similarity is prioritized in sparse graphs, while node feature similarity is emphasized in dense graphs to evaluate hidden edges.
Next, we select the Bottom-$K$ \(\gamma\) values (weakest connections) and set the corresponding edges in \(\textbf{A}_s\) to \(\omega\), obtaining \(\textbf{A}_s^{drop}\):
\begin{equation}
	\textbf{A}_{s,\, i,\, j} = \omega ,
\end{equation} 
where \(\omega\) is a low-intensity value, and $K$ for virtual node $i$ is defined as $ \left \lfloor |\mathcal N(i)|\times \beta\right \rfloor$, with $\beta$ being the edge dropping probability $0<\beta<1$. Here, $i$ and $j$ represent the indices of the two nodes in the Bottom-$K$ \(\gamma\).
This approach balances fine-grained temporal pattern similarity and regional pattern similarity between nodes, optimizing edge selection and enhancing graph quality. 
Furthermore, to allow DarkFarseer to apply SGDs adaptively, the SGDs module is activated only when \(\mathcal{D}>1\).

\subsection{Complexity Analysis}
In summary, DarkFarseer demonstrates good scalability \textit{as its time and space complexities grow linearly rather than exponentially with graph size $N$}.

\paragraph{Time Complexity.}
For \textsc{SeTS}, the time complexity of Equation 7 is $\mathcal O(p)$, and Equation 8 is implemented using a mask, resulting in a time complexity of $\mathcal O(1)$. Therefore, the total time complexity of \textsc{SeTS} is $\mathcal O(p)+\mathcal O(1)=\mathcal O(p)$. 
During the training phase, for \textsc{RsCL},
calculating the prototypes for positive and negative samples for each virtual node incurs a cost of $\mathcal O(mN_u)$, assuming each BCC is utilized. Computing the $\mathcal{L}_{RS}$ for $N_u$ virtual nodes requires $\mathcal O(N_u)$. Therefore, the overall time complexity of \textsc{RsCL} is $\mathcal O(mN_u)+\mathcal O(N_u)=\mathcal O(mN_u)$. 
For SGDs, assuming that the average degree of nodes is $d$ and ignoring the complexity of calculating the Bottom-K, it requires $\mathcal O(dN_u)$.
In summary, disregarding the GNN, the time complexities of DarkFarseer during the training and inference phases are $\mathcal O(p+(m+d)N_u)$ and $\mathcal O(p+dN_u)$, respectively.

\paragraph{Space Complexity.}
For DarkFarseer, a batch of sequences is input with the shape $[B, \varphi, N', p]$, where $B$ represents the batch size and $N'$ is the number of neighboring nodes corresponding to the virtual nodes. \textsc{SeTS} utilizes DLinear (Equation 7) consuming memory $\mathcal{O}(B\varphi N' p)$. 
In \textsc{RsCL} and SGDs, obtaining the prototype of a BCC using mean pooling incurs a cost of $\mathcal O(B\varphi p)$. In \textsc{RsCL}, computing the InfoNCE loss requires $\mathcal O(B\varphi p)$ in Equation 11. In SGDs, calculating $\gamma$ requires $\mathcal O(B\varphi p)$ in Equation 14.
Therefore, disregarding the GNN, the overall space complexity of DarkFarseer during the training and inference phases is $\mathcal{O}(B\varphi N' p+B\varphi p)$.

\section{Experiments and Results}\label{sct:section_5}
\begin{table*}[htbp]
\resizebox{1\linewidth}{!}{
\begin{tabular}{c|c|cccccccc}
% \toprule
\toprule
Datasets                   & Metrics & \multicolumn{1}{c}{MEAN}      & \multicolumn{1}{c}{KNN} & \multicolumn{1}{c}{IGNNK} & \multicolumn{1}{c}{DualSTN} & \multicolumn{1}{c}{INCREASE} & \multicolumn{1}{c}{KCP} & \multicolumn{1}{c}{KITS} & \multicolumn{1}{c}{\textbf{DarkFarseer}} \\ \midrule
                           & MAE     & 79.78                         & 94.804                  & \underline{62.028$\pm$1.076}              & 71.819$\pm$1.039                & 77.238$\pm$0.554                 & 79.485$\pm$0.060            & 86.814$\pm$1.635             & \textbf{59.344$\pm$0.845}\\
                           & RMSE    & 101.791                       & 114.96                  & \underline{91.228$\pm$0.392}              & 106.941$\pm$1.309               & 106.917$\pm$0.122                & 107.688$\pm$0.125           & 129.958$\pm$3.883            & \textbf{86.561$\pm$2.459}                    \\
\multirow{-3}{*}{PEMS03}   & MRE     & 0.457                         & 0.977                   & \underline{0.371$\pm$0.006}               & 0.426$\pm$0.006                 & 0.460$\pm$0.004                  & 0.476$\pm$0.001             & 0.505$\pm$0.010              & \textbf{0.355$\pm$0.005}                     \\
\midrule
                           & MAE     & 89.241                        & 91.036                  & \underline{70.424$\pm$0.670}              & 70.794$\pm$0.538                & 83.344$\pm$0.911                 & 88.114$\pm$3.892            & 72.588$\pm$0.845             & \textbf{67.994$\pm$0.570}                    \\
                           & RMSE    & 116.288                       & 110.426                 & \underline{97.934$\pm$0.976}              & 100.378$\pm$0.630               & 116.856$\pm$0.431                & 117.769$\pm$1.528           & 109.299$\pm$1.304            & \textbf{96.892$\pm$0.520}                    \\
\multirow{-3}{*}{PEMS04}   & MRE     & 0.406                         & 0.443                   & 0.335$\pm$0.003               & 0.333$\pm$0.003                 & 0.395$\pm$0.005                  & 0.419$\pm$0.018             & \textbf{0.319$\pm$0.004}              & \underline{0.323$\pm$0.002}                     \\
\midrule
                           & MAE     & 4.694 & 4.947                   & \underline{3.842$\pm$0.011}               & 4.999$\pm$0.007                 & 3.857$\pm$0.027                  & N/A                     & 4.042$\pm$0.185              & \textbf{3.796$\pm$0.044}                     \\
                           & RMSE    & 7.882 & 8.07                    & \underline{6.420$\pm$0.014}               & 8.202$\pm$0.002                 & 6.902$\pm$0.103                  & N/A                     & 7.144$\pm$0.300              & \textbf{6.403$\pm$0.038}                     \\
\multirow{-3}{*}{PEMS-BAY} & MRE     & 0.075 & 0.079                   & \underline{0.061$\pm$0.000}               & 0.079$\pm$0.000                 & \underline{0.061$\pm$0.000}                  & N/A                     & 0.064$\pm$0.003              & \textbf{0.060$\pm$0.000}                     \\
\midrule
                           & MAE     & 19.499                        & 20.696                  & 18.860$\pm$0.310              & 21.015$\pm$0.736                & 21.984$\pm$0.110                 & 22.979$\pm$0.090            & \underline{18.151$\pm$0.029}             & \textbf{17.555$\pm$0.361}                    \\
                           & RMSE    & 31.902                        & 32.256                  & \underline{29.105$\pm$0.320}              & 33.322$\pm$0.748                & 33.527$\pm$0.188                 & 35.136$\pm$0.174            & 37.814$\pm$0.069             & \textbf{27.933$\pm$0.351}                    \\
\multirow{-3}{*}{AIR36}    & MRE     & 0.229                         & 0.277                   & \underline{0.227$\pm$0.003}               & 0.252$\pm$0.009                 & 0.265$\pm$0.001                  & 0.278$\pm$0.001             & 0.238$\pm$0.001              & \textbf{0.211$\pm$0.004}                     \\
\midrule
                           & MAE     & 4.776                         & 4.912                   & 3.713$\pm$0.031               & 3.998$\pm$0.041                 & \underline{3.205$\pm$0.077}                  & 3.887$\pm$0.001             & 4.778$\pm$0.004              & \textbf{3.199$\pm$0.053}                     \\
                           & RMSE    & 6.557                         & 6.493                   & \underline{5.339$\pm$0.029}               & 5.703$\pm$0.016                 & 5.406$\pm$0.147                  & 6.014$\pm$0.001             & 6.874$\pm$0.016              & \textbf{4.683$\pm$0.048}                     \\
\multirow{-3}{*}{USHCN}    & MRE     & 0.664                         & 0.822                   & 0.488$\pm$0.004               & 0.529$\pm$0.006                 & \underline{0.422$\pm$0.010}                  & 0.511$\pm$0.000             & 0.659$\pm$0.001              & \textbf{0.421$\pm$0.007}                     \\ 
\bottomrule
% \bottomrule

\end{tabular}
	}
    
    % \end{center}
    \caption{Performance comparison of ISK over five datasets, with the results reported as mean{\footnotesize$\pm$std}. The top and second-best results are emphasized in \textbf{bold} and \underline{underlined text}. 
    N/A denotes that augmentation~\cite{li2024non} cannot work on PEMS-BAY.
    }
    \label{tab:compare}%
\end{table*}

% \subsection{Experiment Setup}

\begin{table*}[htbp]
\resizebox{1\linewidth}{!}{
\begin{tabular}{l|ccc|ccc|ccc}
% \toprule
\toprule
% \rowcolor{gray!20}
\multirow{2}{*}{Variants} & \multicolumn{3}{c|}{PEMS04} & \multicolumn{3}{c|}{PEMS-BAY} & \multicolumn{3}{c}{AIR36} \\
% \rowcolor{gray!20}
\cmidrule(r){2-4} \cmidrule(r){5-7} \cmidrule(r){8-10}
 & MAE & RMSE & MRE & MAE & RMSE & MRE & MAE & RMSE & MRE \\
\midrule
w/o \textsc{SeTS}     & 74.594$\pm$0.591 & 105.669$\pm$1.592 & 0.355$\pm$0.002 & 3.806$\pm$0.109 & \textbf{6.385$\pm$0.114} & 0.060$\pm$0.001 & 18.759$\pm$0.544 & 29.288$\pm$0.459 & 0.226$\pm$0.006 \\
w/o TF       & 69.182$\pm$0.529 & 98.496$\pm$0.464 & 0.329$\pm$0.002 & 3.921$\pm$0.060 & 6.445$\pm$0.078 & 0.062$\pm$0.000 & 18.004$\pm$0.651 & 28.377$\pm$0.693 & 0.216$\pm$0.007 \\
w/o \textsc{RsCL}     & 68.244$\pm$0.467 & 97.889$\pm$0.916 & 0.325$\pm$0.002 & 3.827$\pm$0.016 & 6.417$\pm$0.053 & 0.061$\pm$0.000 & 18.125$\pm$0.911 & 28.568$\pm$0.417 & 0.218$\pm$0.010 \\
w/o SGDs     & -              & -              & -             & 3.889$\pm$0.105 & 6.472$\pm$0.036 & 0.062$\pm$0.001 & 18.572$\pm$1.419 & 29.149$\pm$1.108 & 0.223$\pm$0.017 \\
\textbf{DarkFarseer}  & \textbf{67.994$\pm$0.570} & \textbf{96.892$\pm$0.520} & \textbf{0.323$\pm$0.002} & \textbf{3.796$\pm$0.044} & 6.403$\pm$0.038 & \textbf{0.060$\pm$0.000} & \textbf{17.555$\pm$0.361} & \textbf{27.933$\pm$0.351} & \textbf{0.211$\pm$0.004} \\
\bottomrule
% \bottomrule
\end{tabular}
}
\caption{Ablation studies on PEMS04, PEMS-BAY and AIR36.}
\label{tab:ablation}
% \vspace{-4mm}
\end{table*}

\subsubsection{Datasets.} 
We select five datasets from different domains (transportation, air quality, and precipitation), varying in the number of nodes and time steps: PEMS03, PEMS04, PEMS-BAY, AIR-36, and USHCN, to validate the generalization ability of the model. Among them, PEMS03 and PEMS04 belong to sparse PCGs, PEMS-BAY is a dense PCG, while AIR-36 and USHCN are SPGs.

\paragraph{Baselines.} We select two traditional methods: MEAN and KNN, as well as five deep learning-based baselines: IGNNK~\cite{wu2021inductive} (AAAI'21), DualSTN~\cite{hu2023decoupling} (TNNLS'23), INCREASE~\cite{zheng2023increase} (WWW'23), KCP~\cite{li2024non} (AISTATS'24) and KITS~\cite{xu2023kits} (AAAI'25).

\subsubsection{Implementation Details.}
    We set the random seed to 0 for selecting virtual nodes with an observed-to-virtual nodes ratio of 3:1. The dataset is split into training, validation, and test sets in a ratio of 6:2:2. Experiments are conducted four times, and the averaged results are reported. The time window $p$ is set to 24. We normalize the training, validation, and test sets using the global statistics of the observed nodes from the training set. 
    The adjacency matrices constructed for all datasets are normalized using Equation \ref{eq:SPGs}, where $\varepsilon$ is set to $+\infty$ for PEMS03, PEMS04 and PEMS-BAY, and to 0.1 for AIR36 and USHCN. We use Mean Absolute Error (MAE), Root Mean Square Error (RMSE), and Mean Relative Error (MRE) as evaluation metrics. Lower values are better. The code and appendix are available at https://github.com/DKRG-HEU/DarkFarseer.

\begin{figure*}[htbp]
	\centering
	\includegraphics[width=0.8\linewidth]{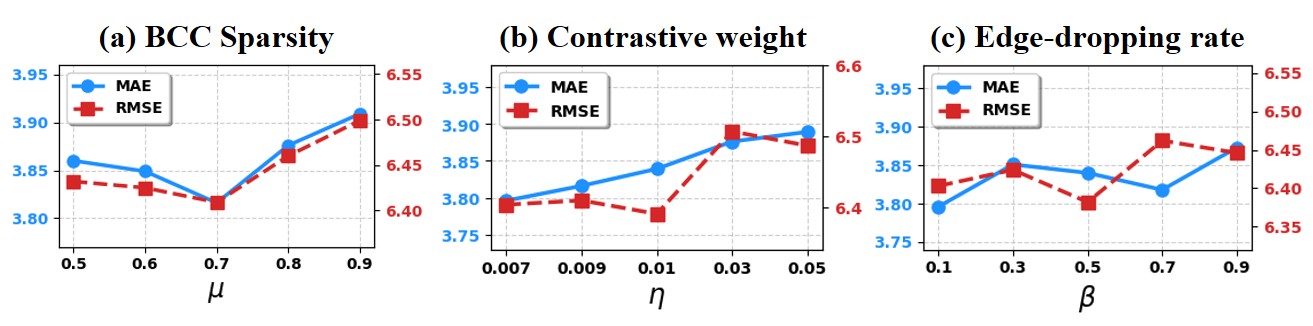} 
	\caption{Impact of BCC sparsity levels $\mu$, contrastive loss weight $\eta$, and edge-dropping rates $\beta$ on PEMS-BAY.
    }
	\label{fig:parameter_study}
\end{figure*}
\begin{figure}[htbp]
	\centering
    % \vspace{-5mm}
	\includegraphics[width=1\linewidth]{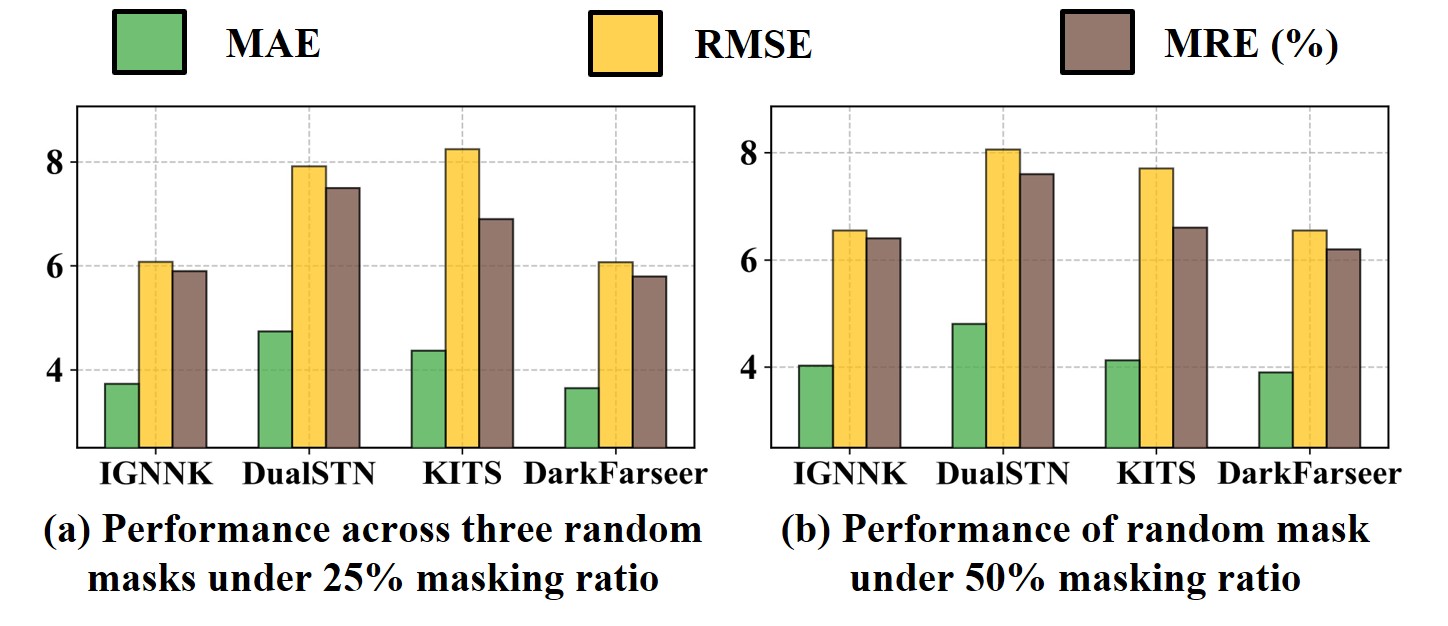} 
	% \vspace{-7mm}
	\caption{
    Evaluation in various scenarios on PEMS-BAY. 
    }
	% \vspace{-6.2mm}
	\label{fig:mask}
\end{figure}
\paragraph{Performance Comparison.}
We compare DarkFarseer with seven baseline methods across five real-world datasets spanning multiple domains. As shown in Table~\ref{tab:compare}, overall, DarkFarseer achieves the best performance. These results underscore the model’s strong generalization.
On \textbf{graph-sparse datasets} (PEMS03 and PEMS04), DarkFarseer consistently outperforms all baselines, including IGNNK. This performance boost is largely attributed to the \textsc{RsCL}, which enriches virtual node representations by contrasting them against regional graph components, thereby alleviating the limitations imposed by sparse graph connectivity. 
On \textbf{graph-dense dataset} (PEMS-BAY), DarkFarseer again achieves SOTA. The \textsc{SGDs} plays a key role here by effectively reducing graph noise, enabling more meaningful message-passing. 
For \textbf{SPG datasets} (AIR36 and USHCN), where noise arises from geographic proximity, DarkFarseer maintains superior performance. The denoising capability of \textsc{SGDs} proves especially valuable in these settings. 
Notably, USHCN, with 1,218 nodes, is a very large-scale graph, and DarkFarseer’s strong performance on it shows its scalability.

\paragraph{Ablation Study.}
We perform an ablation study to assess each key component in DarkFarseer. Specifically, we create the following variants:
\textbf{w/o \textsc{SeTS}}: Removes the \textsc{SeTS} architecture.
\textbf{w/o TF}: Replaces the temporal-first structure with a spatial-first one
(cf. \textbf{L1}).
\textbf{w/o \textsc{RsCL}}: Removes the \textsc{RsCL}.
\textbf{w/o SGDs}: Removes the SGDs.
The results on the PEMS04, PEMS-BAY, and AIR36 datasets, presented in Table 2, demonstrate the importance of each component.
Note that SGDs is not activated on PEMS04.

\paragraph{Robustness Against Diverse Masking Scenarios.}
By default, we focus on random masking with a 25\% missing rate–a prevalent evaluation protocol~\cite{li2024non, li2025feature} that involves proportionally selecting virtual nodes at random.
Here, we set up different masking scenarios:
% \begin{enumerate}[leftmargin=10pt]
\textbf{(1)} We validate the robustness under a 25\% masking ratio by selecting three mask distributions.
\textbf{(2)} We conduct validation under a higher masking ratio (50\%). Masking ratios above 50\% are not considered, as the performance degrades dramatically beyond this point.

We conducted experiments on PEMS-BAY dataset with multiple baselines. The overall results are reported in Figure \ref{fig:mask}.
As shown in Figure \ref{fig:mask}(a), which aggregates results from multiple random mask distributions, DarkFarseer consistently achieves the best performance. This demonstrates the strong generalization of our method.
Furthermore, Figure \ref{fig:mask}(b) presents the performance under a 50\% masking ratio. At higher masking ratios, models must better utilize observed nodes to reconstruct virtual ones. In this challenging context, DarkFarseer still achieves optimal performance.

\paragraph{Parameter Study.}
We conducted parameter studies on the key parameters of \textsc{RsCL} and SGDs using PEMS-BAY.
We analyze the impact of each parameter as follows:
\textbf{BCC sparsity $\mu$:}
The $\mu$ determines sparsity of BCCs, and its value affects the \textsc{RsCL}.
Figure \ref{fig:parameter_study}(a) presents the results concerning $\mu$.
DarkFarseer's performance shows low sensitivity to variations in $\mu$. The optimal performance is observed at $\mu=0.7$, with both excessively large and small $\mu$ values leading to suboptimal results. This suggests that for dense graphs (even with noise), an intermediate BCC sparsity is preferred over extremely sparse ones, indicating a need for balanced topological information rather than maximal dispersion.
\textbf{\textsc{RsCL} weight $\eta$:}
$\eta$ determines the weight of \textsc{RsCL} in the loss. We perform a grid search for $\eta$ within a reasonable range.
Notably, we do not strictly select $\eta$ with equal spacing.
Figure \ref{fig:parameter_study}(b) reports the results. Overall, as $\eta$ varies, DarkFarseer's MAE and RMSE fluctuate minimally. This indicates that DarkFarseer is insensitive to the selection of $\eta$.
\textbf{Edge-dropping rate $\beta$:}
$\beta$ determines the proportion of edges from the virtual node that are downgraded in SGDs. 
Figure \ref{fig:parameter_study}(c) reports the results. Overall, it can be observed that as $\beta$ increases, both MAE and RMSE show a significant upward trend. This aligns with the intuition that neglecting too many edges leads to a decline in the performance of DarkFarseer, which relies on GNNs.

\section{Related Work}\label{sct:section_6}
Kriging is a prominent geostatistical method used for spatial interpolation.
ISK has recently sparked a wave of research~\cite{appleby2020kriging, zhang2022speckriging, hu2023decoupling, xu2023kits, li2024non}.
Deep learning-based ISK methods can be broadly categorized into two paradigms according to the scale of transfer: \textit{partial transfer} and \textit{global transfer}. 
Global transfer aims to migrate the temporal patterns of all nodes to virtual nodes. The most representative approaches employ GNNs. For example, 
\cite{li2024non} utilizes GraphSAGE~\cite{hamilton2017inductive}. Currently, the vast majority of methods rely on global transfer.
Partial transfer generally involves migrating based on the k-nearest neighbors of virtual nodes. 
For instance, \cite{zheng2023increase} aggregates information from heterogeneous k-neighbors.
Nevertheless, \textit{the quality of the graph structure significantly affects the effectiveness of graph-based ISK, yet few works have addressed this issue}.

\section{Conclusion}\label{sct:section_7}
In this study, we tackle key challenges hindering GNN-based ISK: generic network designs that are ill-suited for virtual node inference, and poor graph quality stemming from sparsity and noise. To this end, we propose DarkFarseer, which pioneers a temporal-first structure through \textsc{SeTS}, using style transfer to capture spatio-temporal dependencies around virtual nodes. It further addresses graph quality issues with two innovations: \textsc{RsCL} leverages regional patterns to enrich representations, while the SGDs prunes noisy connections. Extensive experiments on multiple datasets show DarkFarseer achieves SOTA performance.

\section{Acknowledgments}
This research was supported by Natural Science Foundation of Heilongjiang Province, grant number LH2023F020, Supporting Fund of Intelligent Internet of Things and Crowd Computing, grant number B25029, National Natural Science Foundation of China, grant number 62502113, Project (ZR2025QC640) supported by Shandong Provincial Natural Science Foundation, Hainan Provincial Natural Science Foundation of China, grant number 625QN383, and Project (25-1-1-107-zyyd-jch) supported by Qingdao Municipal Natural Science Foundation.

\bibliography{aaai2026}

\end{document}